\documentclass{article}

\PassOptionsToPackage{numbers,compress}{natbib}
\usepackage[main, final]{neurips_2026}

\usepackage[utf8]{inputenc} % allow utf-8 input
\usepackage[T1]{fontenc}    % use 8-bit T1 fonts
\usepackage{hyperref}       % hyperlinks
\usepackage{url}            % simple URL typesetting
\usepackage{booktabs}       % professional-quality tables
\usepackage{amsfonts}       % blackboard math symbols
\usepackage{nicefrac}       % compact symbols for 1/2, etc.
\usepackage{microtype}      % microtypography
\usepackage{xcolor}         % colors
\usepackage{amsmath}
\usepackage{algorithm}
\usepackage{algorithmic}
\usepackage{graphicx}
\usepackage{booktabs}
\usepackage{multirow}
\usepackage[table]{xcolor}
\usepackage{wrapfig}
\usepackage{adjustbox}

\title{What Should a Streaming Video Model Remember?}

\author{
Haonan Ge$^{1,2}$ \quad
Yiwei Wang$^{2\dagger}$ \quad
Hang Wu$^{2}$ \quad
Yujun Cai$^{3}$
\\
$^{1}$University of California, Santa Barbara \quad
$^{2}$University of California, Merced
\\
$^{3}$The University of Queensland
\\
$^{\dagger}$Corresponding author
\\
\texttt{gehaonan82@gmail.com}
\\
\href{https://haonan-ge.github.io/selectstream/}{\texttt{https://haonan-ge.github.io/selectstream/}}
}

\begin{document}

\maketitle

\begin{abstract}
    Streaming video understanding models must answer queries at any moment during an ongoing stream, using only what they have observed so far and under fixed memory and computation budgets. Existing methods address this by adding memory banks, retrieval modules, or visual token compression to preserve long-range history. However, strong recent-window baselines show that indiscriminate history injection can dilute current-scene perception, suggesting that the key challenge is not whether to use memory, but how to allocate it selectively. We formulate this as budgeted online latent evidence allocation and propose \textbf{SelectStream}, a selective latent-memory framework that keeps the current observation directly visible to a frozen VLM while exposing historical information only through a compact, query-conditioned evidence budget. Three coordinated mechanisms govern when to write, what to preserve, and how to retrieve: surprise-driven adaptive windowing, priority-preserving consolidation, and query-conditioned graph reasoning over a fixed-capacity latent memory graph. Retrieved evidence is calibrated and injected as latent tokens for answer generation, without replaying frames or growing the context with stream length. Experimental results show that SelectStream achieves strong online streaming performance and preserves general video understanding, reaching 82.67\% on StreamingBench, 67.03\% on OVO-Bench, and 74.4\% average accuracy on offline video benchmarks, while outperforming strong recent-window baselines and prior streaming memory methods.
\end{abstract}

\section{Introduction}

Streaming video understanding places a distinct demand on vision-language models. In applications such as egocentric life assistants, autonomous driving, robotics, surveillance, and live video interaction, a model must answer queries at any moment during an ongoing stream, using only what has been observed so far and under fixed memory, context, and computation budgets~\citep{yang2025egolife,zeng2025streamforest,wang2025streambridge,yang2025livestar,ge2025framemind,wu2026camreasoner,liu2026passive}. The central question is not merely whether the model can understand video, but whether it can identify and retain the evidence that matters while processing a continuous stream.

Recent work has responded to this challenge by equipping streaming video models with increasingly sophisticated mechanisms for preserving history. External memory banks store visual summaries of past events, KV-cache retrieval mechanisms fetch relevant history on demand, and visual token compression modules distill long temporal context into compact representations~\citep{qian2024streaming,di2025rekv,zeng2025streamforest,zhang2025flashvstream,yao2025timechat,zhang2026hermes,xie2026fluxmem}. The implicit assumption underlying these efforts is that richer memory leads to stronger streaming understanding.

Recent evidence complicates this picture. On standard streaming benchmarks, simple recent-window baselines can match, and in some cases surpass, systems with elaborate memory modules~\citep{shen2026simplestream}. This suggests that incorporating more history does not automatically improve performance. One explanation is evidence dilution: when a model attends over indiscriminately stored historical context, relevant evidence may be mixed with irrelevant content, weakening current-scene perception and reducing the usefulness of retrieved history.

The lesson is not that memory is unnecessary. A pure recent-window strategy has an obvious failure mode: once evidence leaves the local context, it is no longer recoverable. A model that observed a critical object or event minutes earlier still needs a mechanism to access that information later. The real problem is therefore more subtle. A streaming model must decide when to write new evidence, what to keep when capacity runs out, how to retrieve evidence for a given query, and how much historical context to expose at generation time. Existing methods often instantiate these choices through specific memory or compression heuristics, but they are rarely formulated as a unified evidence-allocation problem across all four dimensions.

Motivated by this view, we introduce \textbf{SelectStream}, a selective latent-memory framework for streaming video understanding. SelectStream formulates streaming memory as \emph{budgeted online latent evidence allocation}. It keeps the current observation directly available to a frozen VLM, while historical information is accessed only through a compact query-conditioned latent evidence budget. This design targets the gap between recent-window baselines, which discard older evidence, and memory-heavy systems, which risk exposing too much irrelevant history.

SelectStream implements this idea with a dynamic latent evidence graph. Projected visual embeddings from a frozen backbone VLM are written into fixed-capacity memory, avoiding raw-frame replay and unprojected visual features. Surprise-driven Adaptive Windowing decides when to write, Latent Visual Memory decides what to preserve through priority-aware consolidation, and Graph Attention Reasoning decides how to retrieve compact query-conditioned evidence subgraphs. Retrieved memory states are calibrated and injected as latent evidence tokens, allowing the frozen VLM to use historical evidence without growing its context with stream length.

The main contributions are:
\begin{itemize}
    \item We formulate selective remembering in streaming video understanding as \emph{budgeted online latent evidence allocation}, where a model must decide when to write memory, what evidence to preserve or consolidate, how to read history for a query, and how much evidence to expose under fixed budgets.
    \item We introduce \textbf{SelectStream}, a dynamic latent-memory architecture that writes projected VLM visual embeddings into a fixed-capacity evidence graph, allocates temporal memory resolution with surprise-driven windowing, and preserves useful history through priority-aware consolidation.
    \item We design a query-conditioned graph reasoning and evidence injection interface that retrieves compact historical evidence and injects calibrated latent evidence tokens into a frozen VLM without replaying raw frames or converting memory into text summaries.
\end{itemize}

\begin{figure*}[t]
    \centering
    \includegraphics[width=\textwidth]{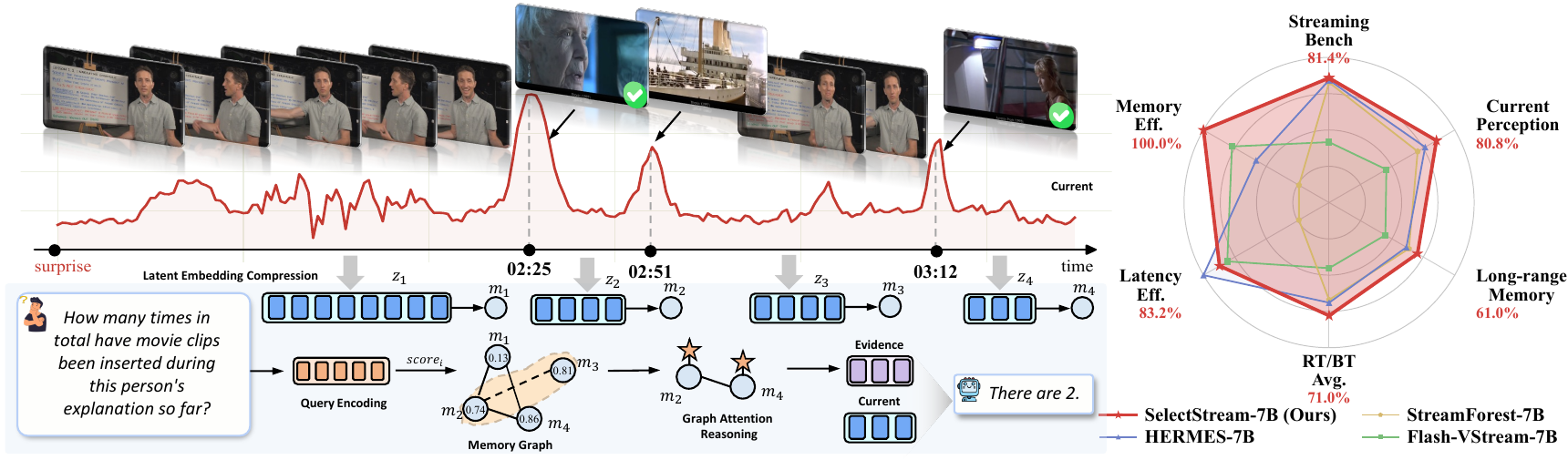}
    \caption{
    Motivation of \textbf{SelectStream}. Relevant events may appear sparsely over a long stream, while later queries require evidence beyond the recent context. The example shows a counting question where inserted movie clips form sparse surprise peaks. The right panel illustrates the desired trade-off: stronger long-range memory and efficiency without sacrificing current-scene perception.
    }
    \label{fig:intro}
\end{figure*}

\section{Related Work}

\vspace{-6pt}
\paragraph{Streaming video understanding.}
Streaming video understanding requires models to answer from a causally observed video prefix rather than a fully available offline video. Recent work studies online video QA, proactive response timing, streaming instruction tuning, and real-time interaction~\citep{qian2024streaming,qian2025dispider,chen2025livecc,xia2025streamo,wang2025streambridge,yang2025livestar,xu2026streamingvlm,ge2025framemind,ge2025mrfd,wu2026camreasoner,xia2025sportr,fu2026videostir,huang2026gui,wu2025refineshot}. These methods address when to respond and how to align perception with generation under causal and computational constraints. SelectStream focuses on preserving, consolidating, and retrieving historical visual evidence under fixed memory and context budgets.

\vspace{-6pt}
\paragraph{Memory and context management.}
Streaming video LLMs must construct a bounded working context from an unbounded stream. Existing methods use memory banks or event memories~\citep{xiong2025streaming,zeng2025streamforest,zhang2025flashvstream,xie2026fluxmem,guo2026eventvstream}, KV-cache memory and retrieval~\citep{di2025rekv,ning2025livevlm,yang2025streammem,zhang2026hermes}, visual token pruning or compression~\citep{yao2025timechat,jin2025streamingassistant,chen2026streamingtom,wang2025hierarchical}, and recurrent or latent states~\citep{qian2024streaming,qian2025dispider,fu2025contextnav}. SelectStream instead organizes history as a dynamic latent evidence graph: projected VLM embeddings are written into memory, consolidated with priority-aware penalties, retrieved as query-conditioned subgraphs, and injected as calibrated latent evidence tokens. A fuller discussion is provided in Appendix~\ref{app:related_work}.

\vspace{-3pt}
\section{Methodology}
\label{sec:method}

\begin{figure*}[t]
    \centering
    \includegraphics[width=\textwidth]{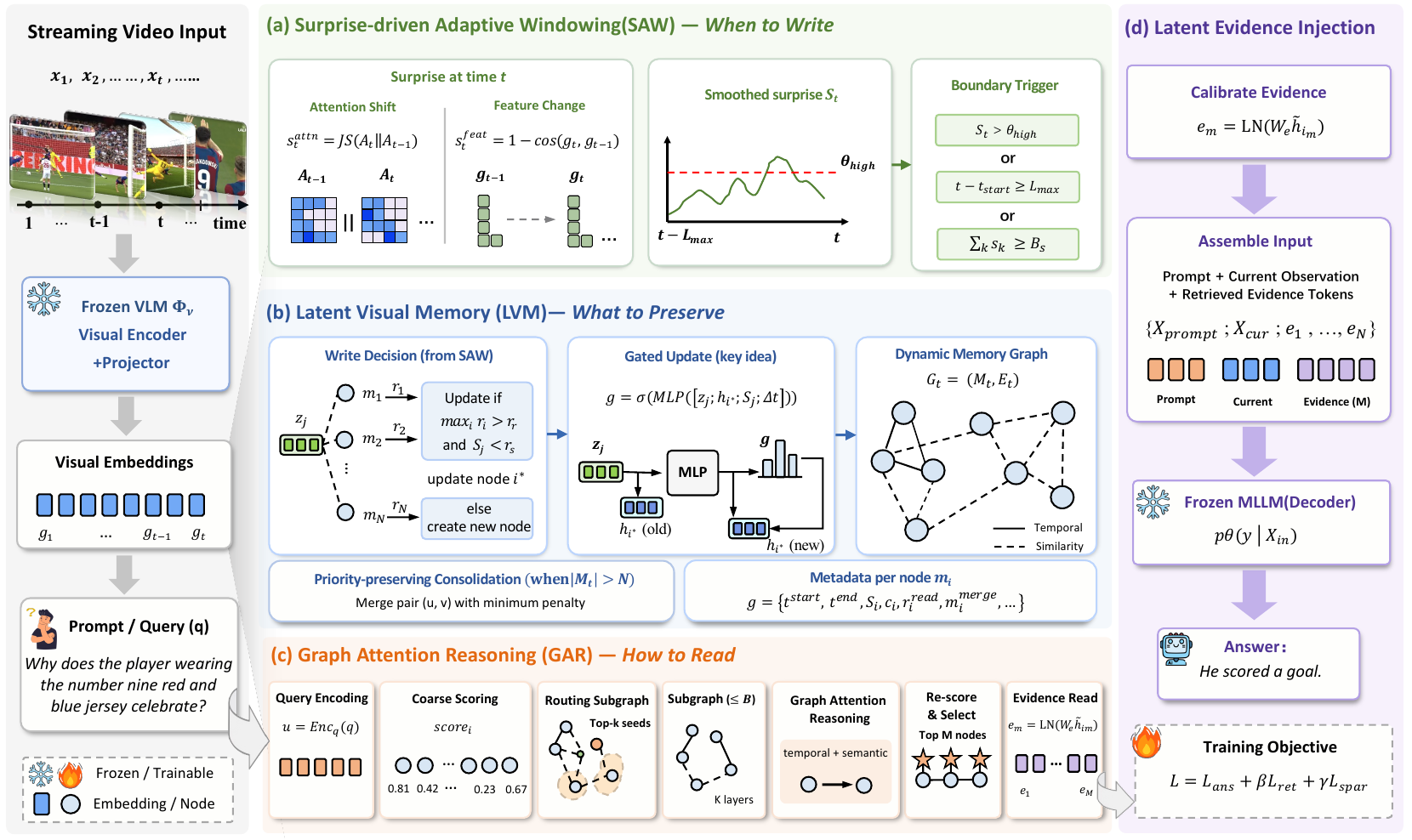}
    \vspace{-12pt}
    \caption{
    Overview of \textbf{SelectStream}. The model writes projected VLM visual embeddings into a budgeted latent memory graph, retrieves a query-conditioned evidence subgraph, and injects calibrated latent evidence tokens into a frozen MLLM for answer generation.
    }
    \label{fig:method}
    \vspace{-9pt}
\end{figure*}

\vspace{-6pt}
SelectStream instantiates the budgeted evidence-allocation view with a dynamic latent evidence graph, as shown in Figure~\ref{fig:method}. It makes three online decisions: Surprise-driven Adaptive Windowing (SAW) decides \emph{when to write}, Latent Visual Memory (LVM) decides \emph{what to preserve} under fixed capacity, and Graph Attention Reasoning (GAR) decides \emph{how to read} compact evidence for a query.

At time step $t$, SelectStream maintains
\begin{equation}
G_t = (\mathcal{M}_t, E_t),
\label{eq:memory_graph}
\end{equation}
where $\mathcal{M}_t$ is the active memory set and $E_t$ contains temporal and semantic relations. Each node stores a latent state $h_i \in \mathbb{R}^d$ and lightweight metadata such as temporal span and surprise statistics. Compared with fixed-window top-$k$ retrieval, SelectStream uses event-adaptive write units, priority-aware consolidation, relation-aware subgraph routing, and calibrated latent evidence injection. The online update loop is summarized in Appendix~\ref{app:online_update_algorithm}.

Unless otherwise specified, SelectStream stores \emph{projected visual embeddings}: each observation is encoded by the frozen VLM visual tower and mapped by its native multimodal projector $\Phi_v(\cdot)$ into the decoder input-embedding space. The primary user-facing budgets are memory capacity $N$, subgraph budget $B$, and evidence budget $M$; SAW also uses $L_{\min}$, $L_{\max}$, and a surprise-energy budget $B_s$ to avoid degenerate segments. Other coefficients are fixed on validation data or learned when applicable, and are reported in Appendix~\ref{app:implementation_hyperparameters}.

\vspace{-3pt}
\subsection{Surprise-driven Adaptive Windowing}
\label{subsec:saw}
\vspace{-3pt}
SAW implements the \emph{when-to-write} decision by assigning coarse memory entries to stable intervals and finer resolution near surprising transitions. For each observation, it caches $g_t=\mathrm{Pool}(\Phi_v(x_t))$ and estimates surprise from attention shift and feature change. By default, $A_t$ is the frozen backbone's prompt-to-visual or visual-token attention, averaged over heads and pooled into fixed visual bins; if such attentions are unavailable, we use a feature-group proxy from projected visual embeddings.

The surprise score is
\begin{equation}
s_t = \lambda s_t^{\text{attn}} + (1-\lambda)s_t^{\text{feat}}, \qquad
\bar{s}_t = \rho \bar{s}_{t-1} + (1-\rho)s_t.
\label{eq:saw_surprise}
\end{equation}
where $s_t^{\text{attn}} = \mathrm{JS}(A_t \,\|\, A_{t-1})$ and $s_t^{\text{feat}} = 1 - \cos(g_t, g_{t-1})$. Head averaging and bin pooling make the JS signal depend on coarse spatial redistribution rather than token-level fluctuations.

A segment closes only after the minimum length is reached and one of three triggers fires: a surprise spike, accumulated surprise energy, or the maximum segment length:
\begin{equation}
t - t_{\mathrm{start}} \ge L_{\min}
\quad \text{and} \quad
\Big(
\bar{s}_t > \theta_{\mathrm{high}}
\;\;\lor\;\;
\sum_{k=t_{\mathrm{start}}}^{t}\bar{s}_k > B_s
\;\;\lor\;\;
t - t_{\mathrm{start}} \ge L_{\max}
\Big),
\label{eq:saw_boundary}
\end{equation}
where $B_s$ is the surprise-energy budget, $L_{\max}$ is the maximum segment length, and $\theta_{\mathrm{high}}$ is an adaptive spike threshold. Proposition~1 formalizes that these triggers control segment generation, while consolidation bounds active memory by $N$. Fixed thresholding and smoothing constants are listed in Appendix~\ref{app:implementation_hyperparameters}.

\subsection{Latent Visual Memory and Dynamic Memory Graph}
\label{subsec:lvm_dmg}
\vspace{-4pt}

LVM implements the \emph{what-to-preserve} decision under fixed capacity. When SAW closes a segment $\mathrm{seg}_j$, cached projected embeddings are encoded as $z_j=\mathrm{SegEnc}(\{g_t:t\in\mathrm{seg}_j\})$ by a lightweight Transformer segment encoder with temporal positions and learned query pooling. Each memory node stores a latent state $h_i \in \mathbb{R}^d$ and metadata such as temporal span, accumulated surprise, write count, read count, and merge count.

To write a segment, we compare $z_j$ with active nodes using $r_i=\cos(z_j,h_i)$. If $\max_i r_i > \tau_r$ and $\bar{s}_j < \tau_s$, the segment updates the selected node $i^*$; otherwise, it creates a new node. The update is gated:
\begin{equation}
g = \sigma\!\left(\mathrm{MLP}\!\left([z_j; h_{i^{*}}; \bar{s}_j; \Delta t]\right)\right), \qquad
h_{i^{*}} \leftarrow (1-g) h_{i^{*}} + g f_{\mathrm{write}}(z_j, h_{i^{*}}).
\label{eq:lvm_write}
\end{equation}
Here $f_{\mathrm{write}}$ is a trainable two-layer MLP over $[z_j;h_{i^*}]$, and $\Delta t$ is the time since the node was last updated. The gate prevents direct replacement from erasing useful history while avoiding unconditional averaging that dilutes important changes.

The graph uses temporal edges $w_{uv}^{\mathrm{tmp}} = \exp(-c \cdot \bar{s}_v)$ for event continuity and similarity edges $w_{uv}^{\mathrm{sim}} = \cos(h_u,h_v)$ for semantic recurrence. When the node budget $N$ is exceeded, SelectStream performs priority-preserving consolidation by merging the pair $(u,v)$ with the smallest graph-aware penalty
\begin{equation}
\pi_{uv}
=
p^{\mathrm{sim}}_{uv}
+
p^{\mathrm{pri}}_{uv}.
\label{eq:merge_penalty}
\end{equation}
The first term favors merging semantically redundant nodes, while the priority term protects surprising, frequently accessed, or recently updated evidence. Full scoring, metadata updates, and distortion derivations are provided in Appendices~\ref{app:consolidation} and~\ref{app:consolidation_error}.

\vspace{-3pt}

\subsection{Query-conditioned Subgraph Retrieval and Graph Attention Reasoning}
\label{subsec:gar}
\vspace{-3pt}

GAR implements the \emph{how-to-read} decision. When a query $q$ arrives, the model routes it through a small memory subgraph rather than reasoning over the full graph. We encode the query as $u = \mathrm{Enc}_q(q)$. In our implementation, $\mathrm{Enc}_q$ appends learned query tokens to the current prompt-side hidden states and applies a lightweight Transformer; the resulting query tokens are mean-pooled into $u$ in the same latent space as memory states.

We first assign each active node a coarse relevance score:
\begin{equation}
\mathrm{score}_i =
\cos(u, h_i)
+ \eta \hat{s}_i
- \xi_{\ell}\hat{\ell}_i
- \xi_m\hat{m}_i^{\mathrm{merge}}.
\label{eq:retrieval_score}
\end{equation}
Here $\ell_i=t_i^{\mathrm{end}}-t_i^{\mathrm{start}}+1$, and $\hat{\ell}_i$ and $\hat{m}_i^{\mathrm{merge}}$ denote active-memory normalized span length and merge count, respectively. The first term measures semantic compatibility, the normalized surprise term $\hat{s}_i$ biases retrieval toward salient events, and the last two terms mildly downweight coarse nodes with long temporal spans or many prior merges. Thus merged nodes can still be retrieved when semantically relevant, but precise grounding prefers sharper unmerged alternatives when available. The scalar coefficients are fixed implementation constants reported in Appendix~\ref{app:implementation_hyperparameters}.

We choose top-$k$ seeds and perform query-conditioned routing over their temporal and semantic neighborhoods. For a routed edge $(i,j)$, the neighbor routing score is
\begin{equation}
\psi_{ij}(q)=\mathrm{score}_j+\alpha_r w_{ij},
\label{eq:routing_score}
\end{equation}
where $w_{ij}$ is the normalized temporal or similarity edge support. If the expansion exceeds the subgraph budget $B$, only the highest-scoring routed nodes are retained. This makes expansion a query-dependent evidence routing step rather than an unconditional hop expansion.

Reasoning is performed on the retrieved subgraph with relational graph attention:
\begin{equation}
e_{ij} = q_i^\top k_j + b_{\mathrm{type}} + b_{\Delta t} + b_w,
\label{eq:gar_logit}
\end{equation}
where $b_{\mathrm{type}}$ is a learned scalar bias for the edge type, $b_{\Delta t}$ penalizes large temporal gaps with a learnable positive scale, and $b_w$ adds edge-support bias. The precise temporal-gap parameterization is given in Appendix~\ref{app:gar_bias}. The normalized attention update is
\begin{equation}
\alpha_{ij} = \mathrm{softmax}_j(e_{ij}), \qquad
h_i^{(\ell+1)} = h_i^{(\ell)} + \sum_{j \in \mathcal{N}(i)} \alpha_{ij} v_j.
\label{eq:gar_update}
\end{equation}
After $K$ layers, each candidate node has gathered temporal and semantic context from its neighborhood. We re-score the refined nodes with the same query-conditioned terms in Eq.~\eqref{eq:retrieval_score}, select the top-$M$ nodes, and read them out as evidence tokens:
\begin{equation}
E=\{e_m\}_{m=1}^{M},\qquad e_m = \mathrm{LN}(W_e \tilde{h}_{i_m}),
\label{eq:evidence_readout}
\end{equation}
where $\tilde{h}_{i_m}$ is a selected refined node state. $W_e$ is a trainable calibration/projection layer followed by LayerNorm into the MLLM token-embedding dimension. Since memory states already originate from projected visual embeddings, $W_e$ is not a full cross-modal translator; it calibrates distributional drift introduced by segment encoding, gated writing, consolidation, and graph reasoning.

\subsection{Latent Evidence Injection for Answer Generation}
\label{subsec:latent_injection}

Latent evidence injection is the final allocation step: only the retrieved evidence budget $M$ is exposed to the MLLM. Instead of replaying frames or converting history into text summaries, the retrieved evidence is injected as latent embeddings. At query time, the MLLM consumes the current prompt and visual observation together with the evidence embeddings,
\begin{equation}
\mathbf{X}_{\mathrm{in}} =
[\mathbf{X}_{\mathrm{prompt}};\mathbf{X}_{\mathrm{cur}};e_1,\ldots,e_M],
\qquad
y\sim p_\theta(y\mid \mathbf{X}_{\mathrm{in}}).
\label{eq:latent_injection}
\end{equation}
In practice, the backbone VLM remains frozen, and $W_e$ is trained as part of the SelectStream evidence calibration interface. This separates online memory writing from query-time reading and grounds generation in explicitly retrieved history. Proposition~2 formalizes how reducing the candidate set to $M$ nodes improves a lower bound on attention to the relevant evidence when retrieval recall is high.

\subsection{Theoretical Properties}
\label{subsec:theory}

The following properties provide two practical intuitions behind SelectStream. Proposition~1 shows that adaptive writing remains budget-controlled even when segment boundaries are generated online. Proposition~2 explains why compact query-conditioned retrieval can reduce evidence dilution: when the relevant evidence is included with reasonable recall, exposing fewer distractors increases the attention mass available to that evidence.

\noindent\textbf{Proposition 1 (Bounded segment and memory complexity).}
Let $S_T$ be the number of SAW segments generated before consolidation over a stream of length $T$, counting the final open segment if it is flushed at the end. Assume the adaptive spike threshold is lower-bounded by $\theta_{\min}>0$, i.e., $\theta_{\mathrm{high}}(t)\ge \theta_{\min}$ for all $t$. Then
\begin{equation}
S_T \le
\frac{\sum_{t=1}^{T}\bar{s}_t}{B_s}
+
\frac{\sum_{t=1}^{T}\bar{s}_t}{\theta_{\min}}
+
\frac{T}{L_{\max}} + 1.
\label{eq:segment_memory_bound}
\end{equation}
After consolidation, the active memory graph always satisfies $|\mathcal{M}_T|\le N$.

\noindent\textit{Proof sketch.}
Partition closed segments by the trigger in Eq.~\eqref{eq:saw_boundary}: accumulated surprise, spike surprise, or maximum length. Disjointness bounds the three counts by the surprise mass and the maximum-window budget. Each write either updates an existing node or creates one node, and consolidation restores $|\mathcal{M}_t|\le N$ whenever the budget is exceeded. The complete proof is provided in Appendix~\ref{app:proof_segment_memory}.

\noindent\textbf{Proposition 2 (Evidence attention concentration).}
Consider query-time attention over a retrieved candidate set of size $M$. Suppose the relevant evidence node $h^*$ is included in this set with probability $r$. When included, its attention logit is at least $\mu_{\mathrm{sig}}$, and each irrelevant candidate logit $\ell$ satisfies $\mathbb{E}[\exp(\ell)\mid h^*\ \mathrm{included}]\le \exp(\sigma^2/2)$. Then the expected attention assigned to the relevant evidence is lower-bounded by
\begin{equation}
\mathbb{E}[\alpha_{h^*}]
\ge
r\cdot
\frac{\exp(\mu_{\mathrm{sig}})}
{\exp(\mu_{\mathrm{sig}})+(M-1)\exp(\sigma^2/2)}.
\label{eq:attention_concentration}
\end{equation}

\noindent\textit{Proof sketch.}
Conditioned on retrieving $h^*$, its softmax numerator is at least $\exp(\mu_{\mathrm{sig}})$. The assumed exponential-moment bound controls the expected contribution of each irrelevant candidate, so the expected distractor mass scales with $M-1$. Multiplying by retrieval recall $r$ gives Eq.~\eqref{eq:attention_concentration}. The assumptions on $r$ and $\mu_{\mathrm{sig}}$ describe retrieval quality and evidence separability rather than architectural guarantees; in practice, they are encouraged by retrieval supervision in Eq.~\eqref{eq:retrieval_loss}, answer loss through the injected evidence tokens, and the evidence calibration layer, and are evaluated empirically with evidence Recall@$M$ and temporal-overlap metrics. The complete proof is provided in Appendix~\ref{app:proof_attention_concentration}.

\subsection{Training Objective}
\label{subsec:objective}

We perform supervised fine-tuning with a frozen backbone VLM. Only the SelectStream modules are updated, using the objective
\begin{equation}
\mathcal{L} = \mathcal{L}_{\mathrm{ans}} + \beta \mathcal{L}_{\mathrm{ret}} + \gamma \mathcal{L}_{\mathrm{spar}}.
\label{eq:training_objective}
\end{equation}
Here $\mathcal{L}_{\mathrm{ans}}$ is autoregressive answer loss, $\mathcal{L}_{\mathrm{ret}}$ trains evidence retrieval when evidence annotations are available, and $\mathcal{L}_{\mathrm{spar}}$ regularizes diffuse routing and redundant evidence. For answer-only data, $\mathcal{L}_{\mathrm{ret}}=0$; these examples train answer generation from retrieved evidence but do not provide direct retrieval grounding. The backbone VLM is not updated; the segment encoder, gate/write MLPs, query encoder, GAR module, and evidence calibration/projection layer are trained by this objective. The exact retrieval and sparsity losses are specified in Appendix~\ref{app:training_details}.

Overall, $N$, $B$, and $M$ separately control memory capacity, retrieval scope, and generation context. The stream still requires one frozen visual pass and native multimodal projection per incoming observation, but segment writing reuses cached projected embeddings and query-time reasoning is controlled by these budgets rather than by full video length. Exact consolidation can use all-pair penalties under small $N$ or top-similarity candidates for real-time deployment. With fixed budgets, the active memory state is $O(1)$ with respect to processed stream length; see Appendix~\ref{app:complexity}.

\begin{table}[t]
\vspace{-6pt}
\centering
\setlength{\abovecaptionskip}{2pt}
\setlength{\belowcaptionskip}{3pt}
\caption{
Main results on StreamingBench and OVO-Bench. RT, BT, and FAR denote Real-Time Visual Perception, Backward Tracing, and Forward Active Responding, respectively. Best results are in bold and second-best results are underlined for each metric. 
}
\label{tab:main_results}
\scriptsize
\setlength{\tabcolsep}{3.2pt}
\renewcommand{\arraystretch}{0.90}

\resizebox{0.98\textwidth}{!}{
\begin{tabular}{lcccccc}
\toprule
\toprule
\multirow{2}{*}{Model}
& \multirow{2}{*}{\#Frames}
& \multirow{2}{*}{StreamingBench}
& \multicolumn{4}{c}{OVO-Bench} \\
\cmidrule(lr){4-7}
& & & RT Avg. & BT Avg. & RT/BT Avg. & FAR / Overall \\
\midrule
\rowcolor{orange!6}
\multicolumn{7}{c}{\textit{Offline Video LLMs}} \\
\midrule
Qwen2.5-VL-7B~\citep{bai2025qwen25vl}
& 1 fps & 73.31 & 59.90 & 44.70 & 52.28 & -- / -- \\
LLaVA-OneVision-7B~\citep{li2025llavaonevision}
& 32 & 71.12 & 64.00 & 43.70 & 53.85 & 50.50 / 52.74 \\
InternVL2-8B~\citep{chen2024internvl}
& 16 & 63.72 & 60.40 & 43.40 & 51.90 & 46.60 / 50.15 \\
LLaVA-Video-7B~\citep{zhang2025llavavideo}
& 64 & -- & 63.50 & 40.40 & 51.95 & 54.82 / 52.91 \\
Qwen2-VL-7B~\citep{wang2024qwen2vl}
& 64 & 69.04 & 60.70 & 48.60 & 54.62 & -- / -- \\
LongVU-7B~\citep{shen2025longvu}
& 1 fps & -- & 57.40 & 39.50 & 48.45 & -- / -- \\

\midrule
\rowcolor{orange!6}
\multicolumn{7}{c}{\textit{Online / Streaming Video LLMs}} \\
\midrule
VideoLLM-online-8B~\citep{chen2024videollmonline}
& 2 fps & 35.99 & 20.80 & 17.70 & 19.26 & -- / -- \\
Flash-VStream-7B~\citep{zhang2025flashvstream}
& 1 fps & 23.23 & 28.40 & 27.40 & 27.90 & 45.09 / 33.61 \\
Dispider-7B~\citep{qian2025dispider}
& 1 fps & 67.63 & 54.60 & 36.10 & 45.35 & 34.72 / 41.78 \\
TimeChat-Online-7B~\citep{yao2025timechat}
& 1 fps & 75.28 & 61.90 & 41.70 & 51.80 & -- / -- \\
StreamForest-7B~\citep{zeng2025streamforest}
& 1 fps & 77.26 & 61.20 & 52.00 & 56.60 & -- / -- \\
Streamo-7B~\citep{xia2025streamo}
& 1 fps & -- & 65.98 & 46.10 & 56.04 & 54.77 / 55.61 \\
Streamo-7B~\citep{xia2025streamo}
& 2 fps & -- & 67.44 & 49.18 & 58.31 & \textbf{56.96} / 57.86 \\
HERMES-7B~\citep{zhang2026hermes}
& 1 fps & 79.44 & 69.00 & 49.40 & 59.20 & -- / -- \\
ThinkStream-7B~\citep{liu2026thinkingstreamingvideo}
& 1 fps & 75.00 & 69.12 & 60.68 & 64.90 & -- / -- \\
Qwen2.5-VL-7B + 4f~\citep{shen2026simplestream}
& 1fps & 78.47 & 78.40 & 51.90 & 65.13 & -- / -- \\
Qwen3-VL-8B + 4f~\citep{shen2026simplestream}
& 1fps & 80.59 & \underline{81.40} & 54.00 & 67.70 & -- / -- \\

\rowcolor{blue!7}
\textbf{\textsc{SelectStream}-Qwen2.5-VL-7B}
& 1 fps & \underline{81.42} & 80.85 & \underline{61.05} & \underline{70.95} & 55.23 / \underline{65.71} \\
\rowcolor{blue!7}
\textbf{\textsc{SelectStream}-Qwen3-VL-8B}
& 1 fps & \textbf{82.67} & \textbf{82.76} & \textbf{62.20} & \textbf{72.48} & \underline{56.13} / \textbf{67.03} \\
\bottomrule
\bottomrule
\end{tabular}
}
\vspace{-8pt}
\end{table}

\section{Experiments}
\label{sec:experiments}
\vspace{-3pt}

\subsection{Experimental Setup}
\label{subsec:exp_setup}
\vspace{-3pt}

\paragraph{Training and implementation.}
We fine-tune SelectStream on Streamo-Instruct-465K~\citep{xia2025streamo}. Timestamped examples provide retrieval supervision through overlap between annotated evidence intervals and memory-node spans, while examples without reliable timestamps use answer loss only; details are in Appendix~\ref{app:latent_evidence_analysis}. The backbone VLM is frozen, and only the SelectStream modules are trained. We train separate modules for Qwen2.5-VL-7B and Qwen3-VL-8B on NVIDIA A100 80GB GPUs. Default budgets and hyperparameters are reported in Appendix~\ref{app:implementation_hyperparameters}.

\vspace{-6pt}
\paragraph{Baselines and metrics.}
We compare with recent-window baselines including SimpleStream~\citep{shen2026simplestream}, streaming video models including Streamo~\citep{xia2025streamo}, StreamForest~\citep{zeng2025streamforest}, Flash-VStream~\citep{zhang2025flashvstream}, and HERMES~\citep{zhang2026hermes}, and simple memory policies such as uniform sampling, FIFO, random eviction, and similarity merging. All methods use the same causal protocol with matched token budgets when possible. We report accuracy, query latency, active memory size, injected evidence tokens, and peak GPU memory under fixed budgets $N$, $B$, and $M$. Detailed benchmark and metric definitions are provided in Appendix~\ref{app:benchmarks_metrics}.

\subsection{Main Results}
\label{subsec:main_results}
\vspace{-3pt}

\paragraph{Online streaming benchmarks.}
As shown in Table~\ref{tab:main_results}, SelectStream consistently outperforms offline video LLMs, online/streaming video LLMs, and strong recent-window baselines on StreamingBench and OVO-Bench. It achieves the best StreamingBench scores among all compared methods, reaching 81.42\% with Qwen2.5-VL-7B and 82.67\% with Qwen3-VL-8B. Compared with the corresponding recent-window baselines, this gives gains of 2.95\% and 2.08\%, indicating that selective latent memory provides additional benefit beyond simply retaining the latest frames. On OVO-Bench, the strongest gains appear on Backward Tracing, which directly evaluates the use of prior visual context. SelectStream improves BT Avg. from 51.90\% to 61.05\% on Qwen2.5-VL-7B and from 54.00\% to 62.20\% on Qwen3-VL-8B, while maintaining strong Real-Time Visual Perception scores of 80.85\% and 82.76\%. This leads to RT/BT averages of 70.95\% and 72.48\%, outperforming prior online models such as StreamForest, Streamo, and ThinkStream. Overall, these results show that SelectStream improves history-dependent reasoning while retaining current-scene perception.

\paragraph{Offline video benchmarks.}
To evaluate generalization beyond streaming-specific benchmarks, we further test SelectStream on VideoMME, MLVU, and MVBench using a causal final-query protocol in Table~\ref{tab:offline_generalization}. SelectStream remains competitive while using a bounded latent evidence budget. With Qwen2.5-VL-7B, it improves the average score from 68.3\% to 70.4\%, with gains of 2.7\% and 2.8\% on VideoMME and MLVU. With Qwen3-VL-8B, it reaches the best overall average of 74.4\%, improving over the frozen backbone by 1.7\%. The larger gains on VideoMME and MLVU suggest that latent memory is most helpful for longer videos requiring temporal evidence, while the smaller gain on MVBench is consistent with its stronger focus on short-range perception and action understanding. These results suggest that SelectStream does not trade offline video understanding for streaming efficiency; instead, it preserves the backbone's general visual reasoning ability while adding useful long-range evidence.

\begin{table}[t]
\vspace{-6pt}
\centering
\setlength{\abovecaptionskip}{2pt}
\setlength{\belowcaptionskip}{3pt}
\caption{
Offline video generalization results on VideoMME, MLVU, and MVBench.
}
\label{tab:offline_generalization}
\scriptsize
\setlength{\tabcolsep}{3.2pt}
\renewcommand{\arraystretch}{0.90}

\resizebox{0.98\textwidth}{!}{
\begin{tabular}{lcccccc}
\toprule
\toprule
Model & Size & \#Frames & VideoMME & MLVU & MVBench & Avg. \\
\midrule
\rowcolor{orange!6}
\multicolumn{7}{c}{\textit{Offline Video LLMs}} \\
\midrule
InternVL2-8B~\citep{chen2024internvl}
& 8B & 64 & 54.0 & 64.0 & 65.8 & 61.3 \\
LongVA-7B~\citep{zhang2024longva}
& 7B & 64 & 52.6 & 56.3 & -- & -- \\
LLaVA-OneVision-7B~\citep{li2025llavaonevision}
& 7B & 32 & 58.2 & 64.7 & 56.7 & 59.9 \\
Qwen2-VL-7B~\citep{wang2024qwen2vl}
& 7B & 64 & 63.3 & -- & 67.0 & -- \\
LongVU-7B~\citep{shen2025longvu}
& 7B & 1 fps & 60.6 & 65.4 & 66.9 & 64.3 \\
LLaVA-Video-7B~\citep{zhang2025llavavideo}
& 7B & 64 & 63.3 & 70.8 & 58.6 & 64.2 \\

\midrule
\rowcolor{orange!6}
\multicolumn{7}{c}{\textit{Online / Streaming Video LLMs}} \\
\midrule
MovieChat-7B~\citep{song2024moviechat}
& 7B & 2048 & 38.2 & 25.8 & 55.1 & 39.7 \\
VideoChat-Online-4B~\citep{huang2025videochatonline}
& 4B & -- & 54.4 & 60.8 & 65.2 & 60.1 \\
Dispider-7B~\citep{qian2025dispider}
& 7B & 1 fps & 57.2 & 61.7 & -- & -- \\
StreamForest-7B~\citep{zeng2025streamforest}
& 7B & 1 fps & 61.4 & 70.0 & 70.2 & 67.2 \\
StreamForest-7B (FT-drive)~\citep{zeng2025streamforest}
& 7B & 1 fps & 61.9 & 69.6 & 68.6 & 66.7 \\
Streamo-7B~\citep{xia2025streamo}
& 7B & 1 fps & 67.9 & -- & 72.3 & -- \\

Qwen2.5-VL-7B~\citep{bai2025qwen25vl}
& 7B & max 768 & 65.1 & 70.2 & 69.6 & 68.3 \\
Qwen3-VL-8B~\citep{bai2025qwen3vl}
& 8B & 2 fps, max 2048 & 71.4 & 78.1 & 68.7 & 72.7 \\
\rowcolor{blue!7}
\textbf{\textsc{SelectStream}-Qwen2.5-VL-7B}
& 7B & 1 fps, max 1024
& \textbf{67.8} {\color{green!45!black}{(+2.7)}}
& \textbf{73.0} {\color{green!45!black}{(+2.8)}}
& \textbf{70.4} {\color{green!45!black}{(+0.8)}}
& \textbf{70.4} {\color{green!45!black}{(+2.1)}} \\
\rowcolor{blue!7}
\textbf{\textsc{SelectStream}-Qwen3-VL-8B}
& 8B & 1 fps, max 1024
& \textbf{73.2} {\color{green!45!black}{(+1.8)}}
& \textbf{80.0} {\color{green!45!black}{(+1.9)}}
& \textbf{69.9} {\color{green!45!black}{(+1.2)}}
& \textbf{74.4} {\color{green!45!black}{(+1.7)}} \\

\bottomrule
\bottomrule
\end{tabular}
}
\vspace{-8pt}
\end{table}

\subsection{Ablation Study}
\label{subsec:ablation}

\paragraph{Effectiveness of memory allocation.}
Using SelectStream-Qwen2.5-VL-7B, Table~\ref{tab:ablation_memory} shows that the full memory allocation strategy outperforms simpler writing and consolidation policies, achieving 81.42\% on StreamingBench, 65.71\% on OVO-Bench, and 73.0\% on MLVU. Replacing SAW with fixed segments reduces the scores to 79.86\%, 64.21\%, and 71.2\%, indicating that event-adaptive segmentation helps preserve informative temporal changes. Removing gated writing also causes a consistent drop, suggesting that direct updates are less effective for balancing old and new evidence. Among consolidation policies, FIFO performs worst because it can discard old but still relevant evidence, while similarity-only merging improves over FIFO but remains below the full model. This shows that priority factors such as surprise, recency, and access frequency help preserve useful evidence under a fixed node budget.

\vspace{-6pt}
\paragraph{Effectiveness of evidence readout.}
Using the same Qwen2.5-VL-7B setting, Table~\ref{tab:ablation_readout} shows that retrieved memory must be routed and calibrated before injection. Replacing GAR with top-$k$ retrieval reduces OVO-Bench from 65.71\% to 63.92\% and MLVU from 73.0\% to 71.1\%. Fixed-hop expansion partially recovers performance, but remains below query-conditioned routing because it does not adapt expansion to the query. The largest drop comes from removing evidence calibration, reducing StreamingBench to 78.96\% and OVO-Bench to 63.08\%; this confirms that memory states are not automatically compatible with the frozen VLM input space. Removing $\mathcal{L}_{\mathrm{ret}}$ weakens grounding, while removing $\mathcal{L}_{\mathrm{spar}}$ gives smaller but consistent drops by allowing redundant evidence.

% Required packages:
% \usepackage{multirow}
% \usepackage[table]{xcolor}

\begin{table}[h]
\centering
\footnotesize
\setlength{\tabcolsep}{2.5pt}
\renewcommand{\arraystretch}{0.95}

\begin{minipage}[t]{0.49\linewidth}
\centering
\caption{Ablation on memory allocation under the same node budget.}
\label{tab:ablation_memory}
\vspace{-0.5em}
\resizebox{\linewidth}{!}{
\begin{tabular}{@{}lccc@{}}
\hline
\textbf{Variant} & \textbf{Streaming} & \textbf{OVO-Bench} & \textbf{MLVU} \\
\hline
Fixed segments w/o SAW & 79.86 & 64.21 & 71.2 \\
w/o gated writing & 80.21 & 64.48 & 71.7 \\
FIFO consolidation & 79.42 & 63.61 & 70.4 \\
Similarity-only merging & 80.03 & 64.03 & 71.0 \\
\rowcolor[HTML]{E8EAFF}
\textbf{Priority consolidation} & \textbf{81.42} & \textbf{65.71} & \textbf{73.0} \\
\hline
\end{tabular}
}
\end{minipage}
\hfill
\begin{minipage}[t]{0.49\linewidth}
\centering
\caption{Ablation on evidence readout under the same evidence budget.}
\label{tab:ablation_readout}
\vspace{-0.5em}
\resizebox{\linewidth}{!}{
\begin{tabular}{@{}lccc@{}}
\hline
\textbf{Model} & \textbf{Streaming} & \textbf{OVO-Bench} & \textbf{MLVU} \\
\hline
Top-$k$ retrieval w/o GAR & 79.71 & 63.92 & 71.1 \\
Fixed-hop expansion & 80.09 & 64.28 & 71.6 \\
w/o evidence calibration & 78.96 & 63.08 & 70.2 \\
w/o $\mathcal{L}_{\mathrm{ret}}$ & 80.27 & 64.42 & 71.8 \\
w/o $\mathcal{L}_{\mathrm{spar}}$ & 80.76 & 64.82 & 72.3 \\
\rowcolor[HTML]{E8EAFF}
\textbf{Full SelectStream} & \textbf{81.42} & \textbf{65.71} & \textbf{73.0} \\
\hline
\end{tabular}
}
\end{minipage}

\vspace{-0.8em}
\end{table}

\begin{wraptable}{r}{0.52\textwidth}
\vspace{-0.8em}
\centering
\footnotesize
\setlength{\tabcolsep}{2.5pt}
\renewcommand{\arraystretch}{0.9}

\begin{adjustbox}{max width=\linewidth}
\begin{tabular}{lccccc}
\toprule
Variant & SB & OVO & MLVU & Recall@$M$ & T-Overlap \\
\midrule
No historical evidence & 78.35 & 62.31 & 69.5 & N/A & N/A \\
Random memory tokens & 78.02 & 62.02 & 69.2 & 0.07 & 0.04 \\
Top-$k$ retrieval w/o GAR & 79.71 & 63.92 & 71.1 & 0.56 & 0.39 \\
Raw states w/o calibration & 78.96 & 63.08 & 70.2 & 0.72 & 0.53 \\
Retrieved-frame replay & 80.84 & 64.88 & 72.4 & 0.72 & 0.53 \\
\rowcolor[HTML]{E8EAFF}
\textbf{Calibrated latent evidence} & \textbf{81.42} & \textbf{65.71} & \textbf{73.0} & \textbf{0.72} & \textbf{0.53} \\
\bottomrule
\end{tabular}
\end{adjustbox}

\caption{Temporal grounding of latent evidence.}
\label{tab:latent_evidence_analysis}
\vspace{-0.8em}
\end{wraptable}

\vspace{-6pt}
\subsection{Latent Evidence Analysis}
\label{subsec:latent_evidence_analysis}
\vspace{-4pt}
SelectStream injects the top-$M$ retrieved memory nodes as calibrated latent evidence tokens, so we evaluate temporal grounding and evidence-interface quality. Table~\ref{tab:latent_evidence_analysis} reports accuracy, Recall@$M$, and T-Overlap on timestamped examples; definitions and retrieval-only diagnostics are in Appendix~\ref{app:latent_evidence_analysis}. Recall@$M$ empirically checks the retrieval-recall term in Proposition~2 (Section~\ref{subsec:theory}). Random memory tokens stay close to no history, with only 0.07 Recall@$M$ and 0.04 T-Overlap. Top-$k$ retrieval improves Recall@$M$ to 0.56, but remains below the full model's 0.72 Recall@$M$ and 0.53 T-Overlap. Raw states use the same retrieved nodes as the full model but are 2.46\%, 2.63\%, and 2.8\% lower on StreamingBench, OVO-Bench, and MLVU, showing that temporal alignment alone is insufficient. Calibrated latent evidence performs best, improving over retrieved-frame replay by 0.58\%, 0.83\%, and 0.6\% while preserving the compact evidence-token interface.

\vspace{-6pt}
\subsection{Budget and Efficiency Analysis}
\label{subsec:budget_efficiency}

We vary memory capacity $N$, subgraph budget $B$, and evidence budget $M$ to evaluate accuracy-efficiency trade-offs. We report time to first token (TTFT) and peak GPU memory as the number of observed frames increases. Since SelectStream maintains a fixed-capacity memory graph and retrieves a bounded subgraph, its query-time cost is controlled by $B$ and $M$ rather than the full stream length.

\begin{figure}[h]
\centering
\vspace{-6pt}
\includegraphics[width=\linewidth]{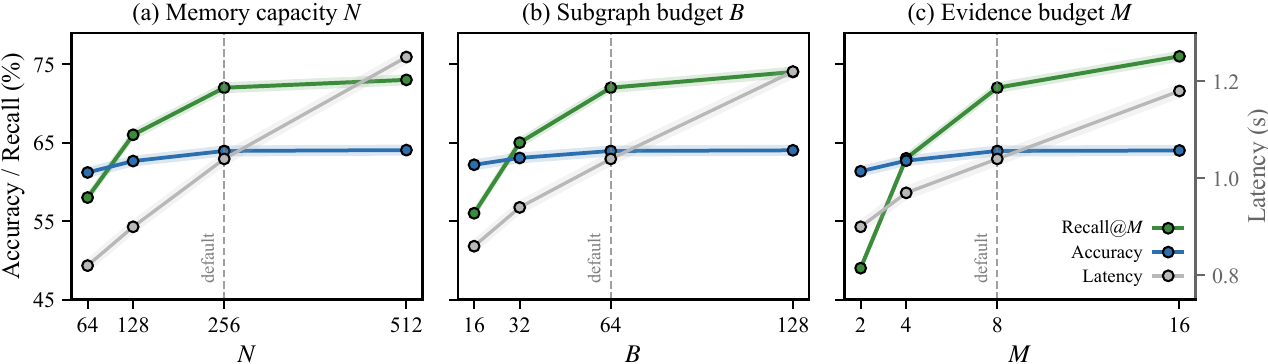}
\vspace{-6pt}
\caption{
Budget sensitivity of SelectStream over memory capacity $N$, subgraph budget $B$, and evidence budget $M$.
}
\label{fig:budget_sensitivity}
\end{figure}

The three exposed budgets show distinct trade-offs in Figure~\ref{fig:budget_sensitivity}. Increasing $N$ improves memory retention, increasing $B$ mainly improves retrieval recall before saturation, and increasing $M$ exposes more evidence but raises TTFT because more latent tokens enter the decoder context. This supports treating $N$, $B$, and $M$ as separate user-facing budgets rather than a single context-length parameter.

\begin{figure}[h]
\centering

\includegraphics[width=\linewidth]{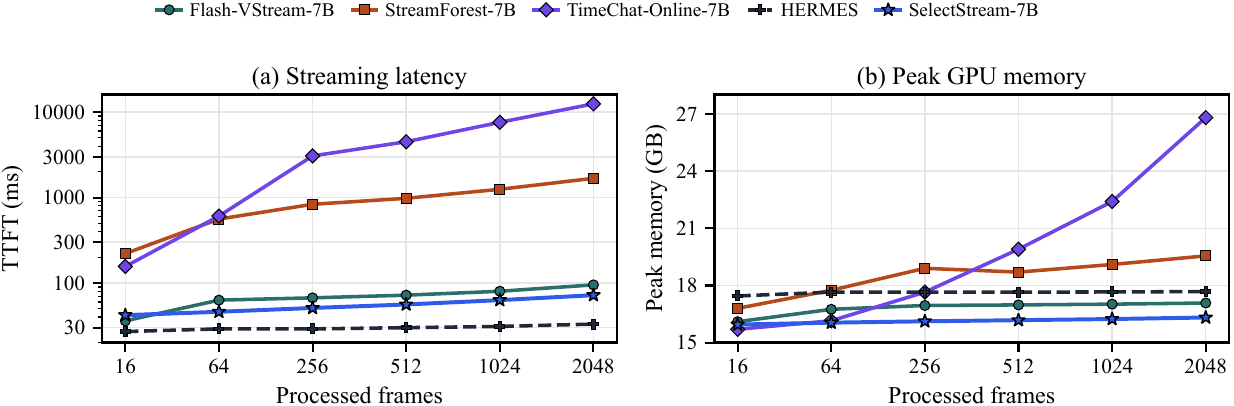}
\vspace{-6pt}
\caption{
Efficiency scaling with stream length. SelectStream keeps query latency and GPU memory nearly flat as processed frames increase.
}

\label{fig:efficiency_scaling}
\end{figure}

The scaling experiment in Figure~\ref{fig:efficiency_scaling} shows that HERMES has the lowest TTFT because its query path is a lightweight KV-cache memory mechanism. SelectStream adds a small startup overhead for node scoring, subgraph routing, GAR re-ranking, and evidence calibration, but its latency grows slowly with more observed frames. Its peak memory also stays nearly flat because the stream is compressed into at most $N$ active nodes and query-time reasoning only materializes a bounded subgraph and $M$ evidence tokens.

\vspace{-8pt}

\vspace{-6pt}
\section{Conclusion}
\label{sec:conclusion}

We presented SelectStream, a streaming video understanding framework that stores long video history as compact latent evidence instead of replaying historical frames. Through event-adaptive writing, priority-aware consolidation, query-conditioned graph retrieval, and calibrated evidence injection, SelectStream improves history-dependent reasoning while preserving current-scene perception and bounded inference cost. Across benchmarks, SelectStream consistently improves long-context video understanding; on OVO-Bench with Qwen2.5-VL-7B, it raises RT/BT Avg. from 65.13\% to 70.95\% and BT Avg. from 51.90\% to 61.05\%. These results support latent evidence allocation as a scalable direction for streaming video understanding under fixed memory and context budgets.

\bibliographystyle{plainnat}
\bibliography{references}
\medskip
\newpage

%%%%%%%%%%%%%%%%%%%%%%%%%%%%%%%%%%%%%%%%%%%%%%%%%%%%%%%%%%%%

\appendix

\section{Additional Related Work}
\label{app:related_work}

\paragraph{Streaming video understanding.}
Streaming video understanding requires models to answer queries from a causally observed video prefix rather than a fully available offline video. Recent work studies this setting through online video question answering, proactive response timing, streaming-oriented instruction tuning, and real-time interaction~\citep{qian2024streaming,qian2025dispider,chen2025livecc,xia2025streamo,wang2025streambridge,yang2025livestar,xu2026streamingvlm}. These works address when to respond, how to align perception with generation under causal constraints, and how to operate under bounded computation. SelectStream focuses on the memory side of this problem: preserving, consolidating, and reading historical visual evidence under fixed memory and context budgets.

\paragraph{Memory and context management for video LLMs.}
A central challenge in streaming video understanding is constructing a bounded working context from an unbounded stream. Existing methods use explicit memory banks or hierarchical event memories~\citep{xiong2025streaming,zeng2025streamforest,zhang2025flashvstream,xie2026fluxmem,guo2026eventvstream}, KV-cache memory and retrieval~\citep{di2025rekv,ning2025livevlm,yang2025streammem,zhang2026hermes}, visual token pruning or compression~\citep{yao2025timechat,jin2025streamingassistant,chen2026streamingtom,wang2025hierarchical}, and learned recurrent or latent states~\citep{qian2024streaming,qian2025dispider}. SelectStream follows this memory-centric line, but organizes history as a dynamic latent evidence graph: projected VLM visual embeddings are written into memory, consolidated with priority-aware penalties, and read as compact query-conditioned subgraphs before calibrated latent evidence tokens are injected into the frozen VLM.

\paragraph{Streaming benchmarks and recency baselines.}
Streaming benchmarks evaluate causal online reasoning rather than offline full-video comprehension. OVO-Bench and StreamingBench test observed-only video understanding with both real-time perception and prior-context use~\citep{li2025ovobench,lin2024streamingbench}, while other benchmarks emphasize proactive assistance, turn-taking, or real-world streaming scenarios~\citep{huang2025ovbench,wang2025omnimmi,wang2025proactivevideoqa,shi2026river,wang2026livibench,lu2026phostream}. Offline long-video benchmarks such as LVBench, MLVU, EgoSchema, and Video-MME evaluate long-range understanding without the same causal constraint~\citep{wang2025lvbench,zhou2025mlvu,mangalam2023egoschema,fu2025videomme}. Recent-window baselines further show that additional memory should be evaluated against simple recency baselines and with disaggregated perception and memory metrics~\citep{shen2026simplestream}. SelectStream is motivated by this standard: current visual evidence remains directly available, while historical memory is exposed only through a compact evidence budget.

\section{Additional Method Details and Proofs}
\label{app:method_details}

This appendix formalizes the implementation details and fixed-budget stability properties of SelectStream, including the online update algorithm, graph-aware memory consolidation, graph-attention bias terms, training losses, complete proofs of the two theoretical properties, default implementation hyperparameters, and a short complexity discussion.

\subsection{Online Update Algorithm}
\label{app:online_update_algorithm}

\begin{algorithm}[h]
\caption{Online visual stream processing and memory update}
\label{alg:streaming_memory}
\begin{algorithmic}[1]
\REQUIRE streaming observations $\{x_1,x_2,\ldots\}$, initialized memory graph $G_0=(\emptyset,\emptyset)$
\REQUIRE memory budget $N$ and fixed SAW/LVM implementation parameters
\STATE Initialize current segment $\mathrm{seg}=[]$ and start time $t_{\mathrm{start}}=1$
\FOR{each observation $x_t$}
    \STATE Extract projected visual embedding $g_t=\mathrm{Pool}(\Phi_v(x_t))$ and compute $\bar{s}_t$ using Eq.~\eqref{eq:saw_surprise}
    \STATE Append cached $g_t$ to $\mathrm{seg}$ and set boundary flag $b_t$ using Eq.~\eqref{eq:saw_boundary}
    \IF{$b_t$}
        \STATE $z_j \leftarrow \mathrm{SegEnc}(\mathrm{seg})$ using cached visual features
        \IF{$\max_i \cos(z_j,h_i)>\tau_r$ and $\bar{s}_j<\tau_s$}
            \STATE Update target node $i^*$ with gated LVM writing in Eq.~\eqref{eq:lvm_write}
        \ELSE
            \STATE Create a new memory node in $G_t$
        \ENDIF
        \STATE Update temporal and similarity edges
        \IF{$|\mathcal{M}_t|>N$}
            \STATE Merge the node pair with minimum penalty $\pi_{uv}$ in Eq.~\eqref{eq:merge_penalty}
        \ENDIF
        \STATE Reset $\mathrm{seg}=[]$ and $t_{\mathrm{start}}=t+1$
    \ENDIF
\ENDFOR
\end{algorithmic}
\end{algorithm}

\subsection{Graph-aware Memory Consolidation}
\label{app:consolidation}

When the active memory size exceeds the budget $N$, the model frees one slot by merging a pair of active nodes. Let $\mathcal{A}_t \subseteq \mathcal{M}_t$ be the active node set. We use the graph-aware merge penalty defined in Eq.~\eqref{eq:merge_penalty}. This section specifies the normalized component scores, the fixed implementation weights, and metadata updates. The pair with the smallest penalty is merged. Each component is normalized to lie in $[0,1]$:
\begin{align}
p^{\mathrm{sim}}_{uv}
&=
\frac{1-\cos(h_u,h_v)}{2},\\
p^{\mathrm{sup}}_{uv}
&=
\frac{\hat{s}_u+\hat{s}_v}{2},\\
p^{\mathrm{acc}}_{uv}
&=
\frac{\hat{r}_u+\hat{r}_v}{2},\\
p^{\mathrm{rec}}_{uv}
&=
\frac{\hat{\tau}_u+\hat{\tau}_v}{2}.
\end{align}
The priority term used in the main text is the fixed weighted combination
\begin{equation}
p^{\mathrm{pri}}_{uv}
=
\lambda_{\mathrm{sup}}
p^{\mathrm{sup}}_{uv}
+
\lambda_{\mathrm{acc}}
p^{\mathrm{acc}}_{uv}
+
\lambda_{\mathrm{rec}}
p^{\mathrm{rec}}_{uv}.
\end{equation}
Equivalently, the implementation supports the explicit form
\begin{equation}
\pi_{uv}
=
\lambda_{\mathrm{sim}}p^{\mathrm{sim}}_{uv}
+
\lambda_{\mathrm{sup}}p^{\mathrm{sup}}_{uv}
+
\lambda_{\mathrm{acc}}p^{\mathrm{acc}}_{uv}
+
\lambda_{\mathrm{rec}}p^{\mathrm{rec}}_{uv},
\end{equation}
with $\lambda_{\mathrm{sim}}=1$ in our default setting, which yields the compact main-text form in Eq.~\eqref{eq:merge_penalty}. Here $\hat{s}_i$ is the normalized accumulated surprise of node $i$, $\hat{r}_i$ is the normalized read count, and $\hat{\tau}_i$ is a normalized recency score computed from the node's latest timestamp. A larger $\hat{\tau}_i$ means the node has been updated more recently. In practice, each statistic is divided by the maximum value among currently active nodes, with a small $\epsilon$ added for numerical stability.

The interpretation is simple. The similarity term is small when two nodes are semantically redundant. The priority term is large for nodes that are likely to be important: surprising nodes may correspond to event boundaries, frequently read nodes are useful for answering past queries, and recently updated nodes may still be temporally active. Minimizing $\pi_{uv}$ therefore prefers merging nodes that are similar, unsurprising, rarely read, and not recently updated, while the main method only exposes the priority-preserving consolidation principle.

After choosing the pair $(u,v)$, the retained node stores a weighted centroid
\begin{equation}
h_{uv}
=
\mathrm{Norm}\!\left(
\frac{w_u h_u+w_v h_v}{w_u+w_v}
\right),
\qquad
w_i=\max(c_i,1),
\end{equation}
where $c_i$ is the write count and $\mathrm{Norm}(\cdot)$ optionally normalizes the state to unit length. The retained metadata is updated by
\begin{align}
t_{uv}^{\mathrm{start}} &= \min(t_u^{\mathrm{start}},t_v^{\mathrm{start}}),\\
t_{uv}^{\mathrm{end}} &= \max(t_u^{\mathrm{end}},t_v^{\mathrm{end}}),\\
\bar{s}_{uv} &= \frac{w_u\bar{s}_u+w_v\bar{s}_v}{w_u+w_v},\\
c_{uv} &= c_u+c_v,\\
r_{uv}^{\mathrm{read}} &= r_u^{\mathrm{read}}+r_v^{\mathrm{read}},\\
m_{uv}^{\mathrm{merge}} &= m_u^{\mathrm{merge}}+m_v^{\mathrm{merge}}+1.
\end{align}
Temporal edges incident to either merged node are inherited by the retained node using max aggregation, and similarity edges are recomputed from the new latent state. The released slot is cleared and becomes available for future writes.

\subsection{Priority-weighted Consolidation Error}
\label{app:consolidation_error}

The consolidation rule should not be interpreted as a globally optimal compression algorithm. Its role is more modest: the similarity term controls feature distortion, while the priority term discourages merging important nodes. The following lemma makes the feature part explicit.

\noindent\textbf{Lemma 1 (Two-node centroid distortion).}
Assume $h_u$ and $h_v$ are $\ell_2$-normalized and let $w_u,w_v>0$. If two nodes are replaced by their weighted centroid
\begin{equation}
\mu_{uv}=\frac{w_u h_u+w_v h_v}{w_u+w_v},
\end{equation}
then the weighted feature distortion introduced before optional renormalization is
\begin{equation}
w_u\|h_u-\mu_{uv}\|_2^2
+
w_v\|h_v-\mu_{uv}\|_2^2
=
\frac{w_u w_v}{w_u+w_v}\|h_u-h_v\|_2^2
=
\frac{2w_u w_v}{w_u+w_v}(1-\cos(h_u,h_v)).
\end{equation}

\noindent\textit{Proof.}
Let $W=w_u+w_v$. Since
\begin{equation}
h_u-\mu_{uv}=\frac{w_v}{W}(h_u-h_v),
\qquad
h_v-\mu_{uv}=-\frac{w_u}{W}(h_u-h_v),
\end{equation}
we have
\begin{align}
w_u\|h_u-\mu_{uv}\|_2^2
+
w_v\|h_v-\mu_{uv}\|_2^2
&=
w_u\frac{w_v^2}{W^2}\|h_u-h_v\|_2^2
+
w_v\frac{w_u^2}{W^2}\|h_u-h_v\|_2^2\\
&=
\frac{w_u w_v(w_u+w_v)}{W^2}\|h_u-h_v\|_2^2\\
&=
\frac{w_u w_v}{w_u+w_v}\|h_u-h_v\|_2^2.
\end{align}
Because $\|h_u\|_2=\|h_v\|_2=1$,
\begin{equation}
\|h_u-h_v\|_2^2
=
\|h_u\|_2^2+\|h_v\|_2^2-2h_u^\top h_v
=
2(1-\cos(h_u,h_v)).
\end{equation}
This proves the claim.

Lemma 1 explains the use of $p^{\mathrm{sim}}_{uv}$ in the merge penalty: for normalized memory states, cosine similarity directly controls the feature distortion of centroid merging. The priority term does not measure feature distortion. Instead, it makes it more costly to merge nodes that are surprising, frequently accessed, or recently updated.

\subsection{Graph Attention Bias Details}
\label{app:gar_bias}

For the GAR attention logit in Eq.~\eqref{eq:gar_logit}, $b_{\mathrm{type}}$ is a learned scalar bias for the edge type (temporal or similarity), $b_{\Delta t}$ is a temporal-gap penalty with a learnable positive scale, and $b_w$ is an additive edge-support term. For nodes with temporal spans $[t_i^s,t_i^e]$ and $[t_j^s,t_j^e]$, we use
\begin{equation}
b_{\Delta t}
=
-\frac{\Delta t_{ij}}{\tau_t},
\qquad
\Delta t_{ij}
=
\max(t_i^s-t_j^e,\;t_j^s-t_i^e,\;0),
\qquad
\tau_t=\mathrm{softplus}(\hat{\tau}_t)+\epsilon.
\label{eq:temporal_gap_bias}
\end{equation}
The edge-support term is $b_w=w_{ij}$, which biases attention toward stronger temporal-continuity or semantic-similarity edges. For routing, edge supports are normalized to $[0,1]$: similarity edges use $(1+\cos(h_i,h_j))/2$, while temporal edges are already non-negative.

\subsection{Training Loss Details}
\label{app:training_details}

Retrieval supervision trains the evidence-allocation policy when evidence annotations are available. Let $\mathcal{P}$ be positive memory nodes: nodes whose spans overlap annotated evidence intervals, or nodes whose IDs match explicit evidence annotations. With retrieval distribution $p_i^{\mathrm{ret}}=\mathrm{softmax}_i(\mathrm{score}_i)$ over active or retrieved candidates, we use a multi-positive loss
\begin{equation}
\mathcal{L}_{\mathrm{ret}}=-\log \sum_{i\in\mathcal{P}} p_i^{\mathrm{ret}},
\label{eq:retrieval_loss}
\end{equation}
when $\mathcal{P}$ is nonempty. Other candidates from the same video act as negatives through the softmax denominator.

The sparsity term penalizes diffuse routing probabilities and redundant evidence:
\begin{equation}
\mathcal{L}_{\mathrm{spar}}
=
\frac{1}{B}\sum_{i\in\mathcal{A}_q} p_i^{\mathrm{route}}
+
\frac{1}{M(M-1)}
\sum_{m\ne n}\max(0,\cos(e_m,e_n)-\delta),
\label{eq:sparsity_loss}
\end{equation}
where $\mathcal{A}_q$ denotes the scored candidate nodes before hard budget pruning, and $\delta$ is a fixed redundancy margin. For the differentiable routing-size proxy, we use
\begin{equation}
p_i^{\mathrm{route}}
=
\sigma\!\left(\frac{\mathrm{score}_i-\kappa}{\tau_{\mathrm{route}}}\right),
\end{equation}
where $\kappa$ and $\tau_{\mathrm{route}}$ are fixed threshold and temperature constants. The hard top-$k$/budgeted routing rule is still used to construct the retrieved subgraph at inference time.

\subsection{Proof of Proposition 1}
\label{app:proof_segment_memory}

Let $S_T$ be the number of SAW segments generated before consolidation over a stream of length $T$, counting the final open segment if it is flushed at the end. Let $\mathcal{C}$ be the set of segment closing times up to time $T$. Each closing time is assigned to one active trigger in Eq.~\eqref{eq:saw_boundary}. If multiple triggers hold simultaneously, we break ties using any fixed order; this only makes the following partition more conservative. Let $\mathcal{C}_B$, $\mathcal{C}_H$, and $\mathcal{C}_L$ denote closures assigned to accumulated surprise, instantaneous high surprise, and maximum length, respectively.

For a closure assigned to accumulated surprise, the corresponding segment interval $I_j$ satisfies
\begin{equation}
\sum_{t\in I_j}\bar{s}_t > B_s.
\end{equation}
The intervals of different closed segments are disjoint. Therefore
\begin{equation}
|\mathcal{C}_B|B_s
\le
\sum_{j:\,t_j\in\mathcal{C}_B}\sum_{t\in I_j}\bar{s}_t
\le
\sum_{t=1}^{T}\bar{s}_t,
\end{equation}
which gives
\begin{equation}
|\mathcal{C}_B|
\le
\frac{\sum_{t=1}^{T}\bar{s}_t}{B_s}.
\end{equation}

For a closure assigned to the high-surprise trigger, the assumption $\theta_{\mathrm{high}}(t)\ge \theta_{\min}>0$ implies
\begin{equation}
\bar{s}_t>\theta_{\mathrm{high}}(t)\ge\theta_{\min}.
\end{equation}
Since closing times are distinct,
\begin{equation}
|\mathcal{C}_H|\theta_{\min}
\le
\sum_{t\in\mathcal{C}_H}\bar{s}_t
\le
\sum_{t=1}^{T}\bar{s}_t,
\end{equation}
and hence
\begin{equation}
|\mathcal{C}_H|
\le
\frac{\sum_{t=1}^{T}\bar{s}_t}{\theta_{\min}}.
\end{equation}

For a closure assigned to the maximum-length trigger, the corresponding segment contains at least $L_{\max}$ time steps under the segment-length convention used by Eq.~\eqref{eq:saw_boundary}. Since such segments are disjoint,
\begin{equation}
|\mathcal{C}_L|L_{\max}\le T,
\qquad
|\mathcal{C}_L|\le \frac{T}{L_{\max}}.
\end{equation}
Adding the three disjoint closure classes and allowing one unfinished open segment at the end gives
\begin{equation}
S_T
\le
\frac{\sum_{t=1}^{T}\bar{s}_t}{B_s}
+
\frac{\sum_{t=1}^{T}\bar{s}_t}{\theta_{\min}}
+
\frac{T}{L_{\max}}
+1.
\end{equation}

It remains to prove the active memory bound. Initially, $|\mathcal{M}_0|=0\le N$. Suppose $|\mathcal{M}_{t-1}|\le N$ before a segment write. The write either updates an existing node, in which case the size does not increase, or creates one new node, in which case the size can become at most $N+1$. If it exceeds $N$, the consolidation step merges one pair of active nodes and releases one slot, returning the size to at most $N$. By induction, $|\mathcal{M}_t|\le N$ for all $t$, and in particular $|\mathcal{M}_T|\le N$.

\subsection{Proof of Proposition 2}
\label{app:proof_attention_concentration}

Let $I$ be the event that the relevant evidence node $h^*$ is included in the retrieved candidate set, with $\Pr(I)=r$. If $I$ does not hold, we set the attention mass assigned to $h^*$ to zero. Conditioned on $I$, let $\ell_*$ be the logit of $h^*$ and let $\ell_1,\ldots,\ell_{M-1}$ be the logits of the irrelevant candidates, where $\mathbb{E}[\exp(\ell_j)\mid I]\le\exp(\sigma^2/2)$ for each distractor. Define
\begin{equation}
a=\exp(\mu_{\mathrm{sig}}),
\qquad
X=\sum_{j=1}^{M-1}\exp(\ell_j).
\end{equation}
By assumption, $\ell_*\ge\mu_{\mathrm{sig}}$, so $\exp(\ell_*)\ge a$. The softmax weight on $h^*$ satisfies
\begin{equation}
\alpha_{h^*}
=
\frac{\exp(\ell_*)}{\exp(\ell_*)+X}
\ge
\frac{a}{a+X}.
\end{equation}
The function $f(X)=a/(a+X)$ is convex for $X\ge0$ because $f''(X)=2a/(a+X)^3>0$. By Jensen's inequality,
\begin{equation}
\mathbb{E}\!\left[\frac{a}{a+X}\mid I\right]
\ge
\frac{a}{a+\mathbb{E}[X\mid I]}.
\end{equation}
Using the exponential moment bound for each irrelevant logit,
\begin{equation}
\mathbb{E}[X\mid I]
=
\sum_{j=1}^{M-1}\mathbb{E}[\exp(\ell_j)\mid I]
\le
(M-1)\exp(\sigma^2/2).
\end{equation}
Therefore
\begin{equation}
\mathbb{E}[\alpha_{h^*}\mid I]
\ge
\frac{\exp(\mu_{\mathrm{sig}})}
{\exp(\mu_{\mathrm{sig}})+(M-1)\exp(\sigma^2/2)}.
\end{equation}
Multiplying by $\Pr(I)=r$ gives
\begin{equation}
\mathbb{E}[\alpha_{h^*}]
\ge
r\cdot
\frac{\exp(\mu_{\mathrm{sig}})}
{\exp(\mu_{\mathrm{sig}})+(M-1)\exp(\sigma^2/2)}.
\end{equation}
This proves the proposition. The bound highlights the role of retrieval: if attention were computed over the full history, the same argument would replace $M$ with the number of historical candidates, increasing the distractor term in the denominator.

\section{More Implementation Details}
\label{app:implementation_hyperparameters}

\paragraph{Training setup.}
We keep the backbone VLM frozen and train only the SelectStream modules, including the segment encoder, write gate, query encoder, GAR module, and evidence calibration layer. For both Qwen2.5-VL-7B and Qwen3-VL-8B, we use bfloat16 precision and train separate backbone-specific modules on 8 NVIDIA A100 80GB GPUs. Unless otherwise stated, we use AdamW with learning rate $2\times 10^{-4}$, batch size $1$, gradient accumulation $4$, and one training epoch. Timestamped examples supervise retrieval with both node-ID and temporal-overlap targets; answer-only examples use only the autoregressive answer loss.

\paragraph{Default implementation hyperparameters.}
The main method exposes the architectural budgets $N$, $B$, and $M$ because they directly control memory capacity, retrieval scope, and generation context. All other coefficients are treated as fixed implementation hyperparameters. Unless otherwise stated, we use the same values across datasets and backbones.

\begin{table}[h]
\centering
\small
\caption{Default SelectStream training and implementation hyperparameters.}
\label{tab:default_hyperparameters}
\begin{tabular}{ll}
\toprule
Hyperparameter & Default value \\
\midrule
optimizer & AdamW \\
learning rate & $2\times 10^{-5}$ \\
random seed & $42$ \\
$N$ memory slots & $256$ \\
$B$ retrieved subgraph budget & $64$ \\
$M$ injected evidence tokens & $8$ \\
$\lambda$ attention-surprise weight & $0.5$ \\
$\rho$ surprise EMA decay & $0.9$ \\
$L_{\min}$, $L_{\max}$ & $8$, $64$ \\
$B_s$ surprise-energy budget & $8.0$ \\
adaptive threshold quantile & $0.9$ \\
SAW recent window & $64$ \\
attention bins & $32$ \\
$\tau_r$, $\tau_s$ & $0.75$, $0.35$ \\
$\lambda_{\mathrm{sim}}$, $\lambda_{\mathrm{sup}}$, $\lambda_{\mathrm{acc}}$, $\lambda_{\mathrm{rec}}$ & $1.0$, $0.5$, $0.25$, $0.25$ \\
top-$k$ seeds & $16$ \\
GAR layers $K$ & $2$ \\
$\eta$, $\xi_{\ell}$, $\xi_m$ & $0.2$, $0.05$, $0.05$ \\
$\alpha_r$ routing edge weight & $0.1$ \\
routing threshold $\kappa$ & $0.0$ \\
routing temperature $\tau_{\mathrm{route}}$ & $1.0$ \\
redundancy margin $\delta$ & $0.0$ \\
$\beta$, $\gamma$ loss weights & $0.1$, $0.05$ \\
ID/time retrieval weights & $1.0$, $1.0$ \\
timestamp tolerance & $0.0$ \\
evidence top-$k$ & $8$ \\
\bottomrule
\end{tabular}
\end{table}

\section{Benchmarks and Metrics}
\label{app:benchmarks_metrics}

\paragraph{Evaluation protocol.}
All online evaluations follow a causal streaming protocol: at a query time, the model can only use the observed video prefix and cannot access future frames. SelectStream processes frames sequentially, updates its latent memory online, and answers with the current observation plus retrieved latent evidence. Unless otherwise stated, SelectStream uses 1 fps input, fixed memory capacity $N$, retrieved subgraph budget $B$, and injected evidence budget $M$. For baselines, we use the reported frame rates and context budgets from the corresponding sources when exact budget matching is not possible.

\paragraph{Online streaming benchmarks.}
StreamingBench evaluates observed-only video understanding across perception, reasoning, and temporal-awareness tasks. We report the official overall accuracy, denoted as StreamingBench or SB in compact tables. OVO-Bench separates online video understanding into Real-Time Visual Perception (RT), Backward Tracing (BT), and Forward Active Responding (FAR). RT measures current-scene perception, BT measures reasoning over previously observed visual context, and FAR measures whether the model responds at appropriate future-oriented moments. We report RT Avg., BT Avg., their mean RT/BT Avg., and FAR / Overall when available:
\begin{equation}
\mathrm{RT/BT\ Avg.}
=
\frac{\mathrm{RT\ Avg.}+\mathrm{BT\ Avg.}}{2}.
\end{equation}
In ablation and compact analysis tables, OVO denotes OVO-Bench Overall.

\paragraph{Offline video benchmarks.}
We evaluate offline generalization on VideoMME, MLVU, and MVBench with a causal final-query protocol. The model observes the video stream sequentially and answers only after the final observation, so the setting still uses SelectStream's online memory construction rather than full-video replay. VideoMME and MVBench are reported with their official accuracy metrics, while MLVU is reported with its official M-Avg score. The offline average is computed over the three dataset-level scores:
\begin{equation}
\mathrm{Avg.}
=
\frac{\mathrm{VideoMME}+\mathrm{MLVU}+\mathrm{MVBench}}{3}.
\end{equation}
Entries marked ``--'' indicate unavailable results from the corresponding source and are excluded from averages unless all three scores are reported.

\paragraph{Ablation and latent-evidence metrics.}
Ablations use SelectStream-Qwen2.5-VL-7B with the same frame rate, frozen backbone, memory capacity, and evidence budget as the full model. We report StreamingBench accuracy, OVO-Bench Overall, and MLVU M-Avg. For latent evidence analysis, Recall@$M$ measures whether at least one of the top-$M$ retrieved memory nodes overlaps the annotated evidence interval, and T-Overlap measures the best fraction of the annotated evidence interval covered by a retrieved node. These metrics evaluate whether the injected latent evidence is temporally grounded and empirically estimate the retrieval-recall condition used by Proposition~2 (section~\ref{subsec:theory}).

\paragraph{Efficiency metrics.}
For efficiency and latency, we report time to first token (TTFT) and peak GPU memory as the number of observed frames increases. TTFT includes query-time memory scoring, subgraph routing, GAR re-ranking, evidence calibration, and the first decoding step. Peak GPU memory is measured during streaming inference and includes the active memory graph and the retrieved evidence used for generation. We also track active memory size and injected evidence tokens to verify that query-time cost is controlled by $N$, $B$, and $M$ rather than by the full stream length.

\section{Additional Latent Evidence Analysis}
\label{app:latent_evidence_analysis}

Each injected evidence token in SelectStream is traceable to a retrieved memory node. For training examples with evidence timestamps, we mark a memory node as positive if its temporal span overlaps the annotated evidence interval with a small tolerance:
\begin{equation}
\mathcal{P}(q)=
\{i \mid [t_i^s,t_i^e]\cap [t_q^s-\epsilon,t_q^e+\epsilon]\neq \emptyset\}.
\end{equation}
We use overlap rather than IoU because memory nodes may represent variable-length or merged segments, and overly strict IoU would penalize coarse nodes that still contain the evidence.

For evaluation, let $R_q^M$ be the top-$M$ retrieved nodes for query $q$ and let $\mathcal{G}_q$ be the set of annotated evidence intervals. Since each memory node $i$ stores a temporal span $I_i=[t_i^s,t_i^e]$, we measure whether the retrieved latent evidence is temporally grounded by

\begin{equation}
\mathrm{Recall@}M
=
\frac{1}{|\mathcal{Q}|}
\sum_{q\in\mathcal{Q}}
\mathbf{1}
\left[
\max_{i\in R_q^M,\,g\in\mathcal{G}_q}
\frac{|I_i\cap g|}{|g|}
>0
\right].
\end{equation}
We also report average temporal overlap,
\begin{equation}
\mathrm{T\mbox{-}Overlap}
=
\frac{1}{|\mathcal{Q}|}
\sum_{q\in\mathcal{Q}}
\max_{i\in R_q^M,\,g\in\mathcal{G}_q}
\frac{|I_i\cap g|}{|g|}.
\end{equation}
The overlap denominator is the ground-truth evidence length rather than the union length. This is intentional: a merged memory node may cover a longer span while still containing the annotated evidence, and IoU would over-penalize such coarse but valid memory.
Empirically, Recall@$M$ estimates the retrieval-recall term $r$ used by Proposition~2 in Section~\ref{subsec:theory}, while T-Overlap measures how tightly the retrieved node spans cover the annotated evidence.

Table~\ref{tab:latent_temporal_grounding_app} isolates temporal grounding under different retrieval variants while keeping the evidence budget fixed. In addition to Recall@$M$ and T-Overlap, we report the median retrieved span length to verify that higher recall does not come simply from selecting overly long memory nodes.

\begin{table}[h]
\centering
% Simulated placeholder values for drafting; replace with measured results before submission.
\caption{
Additional temporal grounding analysis of retrieved latent evidence on timestamped examples. Median span length is measured in seconds.
}
\label{tab:latent_temporal_grounding_app}
\small
\setlength{\tabcolsep}{5pt}
\renewcommand{\arraystretch}{1.08}
\begin{tabular}{lccc}
\toprule
Retrieval Policy & Recall@$M$ & T-Overlap & Median Span \\
\midrule
Initial top-$k$ seeds only & 0.49 & 0.33 & 7.8s \\
Top-$k$ + fixed-hop expansion & 0.61 & 0.43 & 10.6s \\
Top-$k$ + query-conditioned routing & 0.69 & 0.50 & 9.4s \\
Query routing + GAR re-ranking & 0.71 & 0.52 & 9.1s \\
\rowcolor[HTML]{E8EAFF}
\textbf{Full SelectStream} & \textbf{0.72} & \textbf{0.53} & 9.0s \\
\bottomrule
\end{tabular}
\end{table}

For qualitative inspection, we also visualize retrieved evidence on the original video timeline. The figure should show the annotated evidence intervals, active memory-node spans, top-$M$ retrieved nodes, and retrieval scores. This verifies that the latent tokens injected into the VLM can be traced back to concrete video regions.

\section{Complexity Under Fixed Budgets}
\label{app:complexity}

Let $d$ be the memory dimension, $N$ the active memory budget, $B$ the retrieved subgraph budget, $M$ the evidence budget, and $K$ the number of GAR layers. Ignoring the frozen visual pass and native multimodal projection cost, SAW maintains constant-size streaming statistics and adds $O(d)$ work per observation for lightweight feature comparison. When a segment is closed, LVM writing reuses cached projected visual embeddings and compares the segment vector with active memory nodes, requiring $O(Nd)$ time. Naive graph-aware consolidation computes all pair penalties in $O(N^2d)$ time when the memory is full, but this cost is controlled by the fixed budget $N$ and can be reduced with approximate nearest-neighbor candidates or top-$k$ similarity neighborhoods.

At query time, retrieval scores all active nodes in $O(Nd)$ time and then routes the query through a bounded subgraph of size at most $B$. GAR reasoning over the retrieved subgraph costs $O(K(|E_B|d+Bd^2))$, where $E_B$ is the edge set inside the retrieved subgraph. Finally, only $M$ evidence embeddings are injected into the language model. Therefore, after the stream has been compressed into memory, query-time reasoning depends on $N$, $B$, and $M$, rather than on the full video length $T$.

\section{Full Results on OVO-Bench and StreamingBench}
\begin{table*}[h]
\centering
\caption{
Full StreamingBench results. OP, CR, CS, ATP, EU, TR, PR, SU, ACP, and CT denote the official StreamingBench sub-tasks.
}
\label{tab:appendix_streamingbench_full}
\small
\setlength{\tabcolsep}{3.8pt}
\renewcommand{\arraystretch}{1.12}
\resizebox{\textwidth}{!}{
\begin{tabular}{lcc*{11}{c}}
\toprule
\toprule
Model & Size & \#Frames & OP & CR & CS & ATP & EU & TR & PR & SU & ACP & CT & All \\
\midrule
Human & -- & -- & 89.47 & 92.00 & 93.60 & 91.47 & 95.65 & 92.52 & 88.00 & 88.75 & 89.74 & 91.30 & 91.46 \\
\midrule
\rowcolor{orange!6}
\multicolumn{14}{c}{\textit{Proprietary MLLMs}} \\
\midrule
Gemini 1.5 Pro & -- & 1 fps & 79.02 & 80.47 & 83.54 & 79.67 & 80.00 & 84.74 & 77.78 & 64.23 & 71.95 & 48.70 & 75.69 \\
GPT-4o & -- & 64 & 77.11 & 80.47 & 83.91 & 76.47 & 70.19 & 83.80 & 66.67 & 62.19 & 69.12 & 49.22 & 73.28 \\
Claude 3.5 Sonnet & -- & 20 & 73.33 & 80.47 & 84.09 & 82.02 & 75.39 & 79.53 & 61.11 & 61.79 & 69.32 & 43.09 & 72.44 \\
\midrule
\rowcolor{orange!6}
\multicolumn{14}{c}{\textit{Offline Video LLMs}} \\
\midrule
Video-LLaMA2 & 7B & 32 & 55.86 & 55.47 & 57.41 & 58.17 & 52.80 & 43.61 & 39.81 & 42.68 & 45.61 & 35.23 & 49.52 \\
VILA-1.5 & 8B & 14 & 53.68 & 49.22 & 70.98 & 56.86 & 53.42 & 53.89 & 54.63 & 48.78 & 50.14 & 17.62 & 52.32 \\
Video-CCAM & 14B & 96 & 56.40 & 57.81 & 65.30 & 62.75 & 64.60 & 51.40 & 42.59 & 47.97 & 49.58 & 31.61 & 53.96 \\
LongVA & 7B & 128 & 70.03 & 63.28 & 61.20 & 70.92 & 62.73 & 59.50 & 61.11 & 53.66 & 54.67 & 34.72 & 59.96 \\
InternVL2 & 8B & 16 & 68.12 & 60.94 & 69.40 & 77.12 & 67.70 & 62.93 & 59.26 & 53.25 & 54.96 & 56.48 & 63.72 \\
Kangaroo & 7B & 64 & 71.12 & 84.38 & 70.66 & 73.20 & 67.08 & 61.68 & 56.48 & 55.69 & 62.04 & 38.86 & 64.60 \\
LLaVA-NeXT-Video & 32B & 64 & 78.20 & 70.31 & 73.82 & 76.80 & 63.35 & 69.78 & 57.41 & 56.10 & 64.31 & 38.86 & 66.96 \\
MiniCPM-V2.6 & 8B & 32 & 71.93 & 71.09 & 77.92 & 75.82 & 64.60 & 65.73 & 70.37 & 56.10 & 62.32 & 53.37 & 67.44 \\
LLaVA-OneVision & 7B & 32 & 80.38 & 74.22 & 76.03 & 80.72 & 72.67 & 71.65 & 67.59 & 65.45 & 65.72 & 45.08 & 71.12 \\
Qwen2.5-VL & 7B & 1 fps & 78.32 & 80.47 & 78.86 & 80.45 & 76.73 & 78.50 & 79.63 & 63.41 & 66.19 & 53.19 & 73.68 \\
\midrule
\rowcolor{orange!6}
\multicolumn{14}{c}{\textit{Online / Streaming Video LLMs}} \\
\midrule
Flash-VStream & 7B & -- & 25.89 & 43.57 & 24.91 & 23.87 & 27.33 & 13.08 & 18.52 & 25.20 & 23.87 & 48.70 & 23.23 \\
VideoLLM-online & 8B & 2 fps & 39.07 & 40.06 & 34.49 & 31.05 & 45.96 & 32.40 & 31.48 & 34.16 & 42.49 & 27.89 & 35.99 \\
Dispider & 7B & 1 fps & 74.92 & 75.53 & 74.10 & 73.08 & 74.44 & 59.92 & 76.14 & 62.91 & 62.16 & 45.80 & 67.63 \\
StreamForest & 7B & 1 fps & 83.11 & 82.81 & 82.65 & 84.26 & 77.50 & 78.19 & 76.85 & 69.11 & 75.64 & 54.40 & 77.26 \\
Qwen2.5-VL-7B + 4f & 7B & 1fps & -- & -- & -- & -- & -- & -- & -- & -- & -- & -- & 78.47 \\
Qwen3-VL-8B + 4f & 8B & 1fps & -- & -- & -- & -- & -- & -- & -- & -- & -- & -- & 80.59 \\
\midrule
\rowcolor{orange!6}
\multicolumn{14}{c}{\textit{SelectStream Framework}} \\
\midrule
\rowcolor{blue!7}
\textbf{\textsc{SelectStream}-Qwen2.5-VL-7B}
& 7B & 1 fps
& 84.50 & 86.20 & 84.80 & 86.00 & 83.20
& 83.40 & 84.60 & 73.80 & 76.70 & 70.99
& 81.42 \\

\rowcolor{blue!7}
\textbf{\textsc{SelectStream}-Qwen3-VL-8B}
& 8B & 1 fps
& 86.30 & 87.40 & 86.00 & 87.10 & 84.70
& 84.60 & 85.80 & 75.00 & 78.20 & 71.60
& 82.67 \\

\bottomrule
\bottomrule
\end{tabular}
}
\end{table*}

\begin{table*}[h]
\centering
\caption{
Detailed evaluation results on OVO-Bench.
}
\label{tab:appendix_ovobench_full}
\tiny
\setlength{\tabcolsep}{2.3pt}
\renewcommand{\arraystretch}{1.03}
\resizebox{\textwidth}{!}{
\begin{tabular}{l|c|cccccc|c|ccc|c|ccc|c|c}
\hline
\hline
\multirow{2}{*}{Model}
& \multirow{2}{*}{\# Frames}
& \multicolumn{7}{c|}{Real-Time Visual Perception}
& \multicolumn{4}{c|}{Backward Tracing}
& \multicolumn{4}{c|}{Forward Active Responding}
& \multirow{2}{*}{Overall Avg.} \\
\cline{3-17}
& & OCR & ACR & ATR & STU & FPD & OJR & Avg.
& EPM & ASI & HLD & Avg.
& REC & SSR & CRR & Avg.
& \\
\hline
Human & -- & 93.96 & 92.57 & 94.83 & 92.70 & 91.09 & 94.02 & 93.20 & 92.59 & 93.02 & 91.37 & 92.33 & 95.48 & 89.67 & 93.56 & 92.90 & 92.81 \\
\hline
\rowcolor{orange!6}
\multicolumn{18}{c}{\textit{Proprietary MLLMs}} \\
\hline

Gemini 1.5 Pro & 1 fps & 85.91 & 66.97 & 79.31 & 58.43 & 63.37 & 61.96 & 69.32 & 58.59 & 76.35 & 52.64 & 62.54 & 35.53 & 74.24 & 61.67 & 57.15 & 63.00 \\
GPT-4o & 64 & 69.80 & 64.22 & 71.55 & 51.12 & 70.30 & 59.78 & 64.46 & 57.91 & 75.68 & 48.66 & 60.75 & 27.58 & 73.21 & 59.40 & 53.40 & 59.54 \\
\hline
\rowcolor{orange!6}
\multicolumn{18}{c}{\textit{Open-source Offline Video MLLMs}} \\
\hline

Qwen2-VL-72B & 64 & 65.77 & 60.55 & 69.83 & 51.69 & 69.31 & 54.35 & 61.92 & 52.53 & 60.81 & 57.53 & 56.95 & 38.83 & 64.07 & 45.00 & 49.30 & 56.27 \\
LLaVA-Video-7B & 64 & 69.13 & 58.72 & 68.83 & 49.44 & 74.26 & 59.78 & 63.52 & 56.23 & 57.43 & 7.53 & 40.40 & 34.10 & 69.95 & 60.42 & 54.82 & 52.91 \\
LLaVA-OneVision-7B & 64 & 66.44 & 57.80 & 73.28 & 53.37 & 71.29 & 61.96 & 64.02 & 54.21 & 55.41 & 21.51 & 43.71 & 25.64 & 67.09 & 58.75 & 50.50 & 52.74 \\
Qwen2-VL-7B & 64 & 60.40 & 50.46 & 56.03 & 47.19 & 66.34 & 55.43 & 55.98 & 47.81 & 35.48 & 56.08 & 46.46 & 31.66 & 65.82 & 48.75 & 48.74 & 50.39 \\
InternVL2-8B & 64 & 67.11 & 60.55 & 63.79 & 46.07 & 68.32 & 56.52 & 60.39 & 48.15 & 57.43 & 24.73 & 43.44 & 26.50 & 59.14 & 54.14 & 46.60 & 50.15 \\
LongVU-7B & 1 fps & 53.69 & 53.21 & 62.93 & 47.75 & 68.32 & 59.78 & 57.61 & 40.74 & 59.46 & 4.84 & 35.01 & 12.18 & 69.48 & 60.83 & 47.50 & 46.71 \\
\hline
\rowcolor{orange!6}
\multicolumn{18}{c}{\textit{Open-source Online Video MLLMs}} \\
\hline

VideoLLM-online-8B & 2 fps & 8.05 & 23.85 & 12.07 & 14.04 & 45.54 & 21.20 & 20.79 & 22.22 & 18.80 & 12.18 & 17.73 & -- & -- & -- & -- & 12.84 \\
Flash-VStream-7B & 1 fps & 24.16 & 29.36 & 28.45 & 33.71 & 25.74 & 28.80 & 28.37 & 39.06 & 37.16 & 5.91 & 27.38 & 8.02 & 67.25 & 60.00 & 45.09 & 33.61 \\
Dispider-7B & 1 fps & 57.72 & 49.54 & 62.07 & 44.94 & 61.39 & 51.63 & 54.55 & 48.48 & 55.41 & 4.30 & 36.06 & 18.05 & 37.36 & 48.75 & 34.72 & 41.78 \\
ViSpeak-7B & 1 fps & 75.17 & 58.72 & 71.55 & 51.12 & 74.26 & 66.85 & 66.28 & 59.93 & 48.65 & 63.98 & 57.52 & 33.81 & 68.52 & 60.42 & 54.25 & 61.08 \\
StreamForest-7B & 1 fps & 68.46 & 53.21 & 71.55 & 47.75 & 65.35 & 60.87 & 61.20 & 58.92 & 64.86 & 32.26 & 52.02 & 32.81 & 70.59 & 57.08 & 53.49 & 55.57 \\
ET-Instruct-3B & 1 fps & 65.10 & 35.78 & 56.90 & 35.39 & 24.75 & 60.87 & 46.47 & 41.81 & 35.14 & 8.60 & 28.52 & 20.06 & 52.31 & 67.50 & 46.62 & 40.54 \\
ET-Instruct-3B$^\dagger$ & 1 fps & 71.14 & 50.46 & 67.24 & 37.08 & 60.40 & 60.33 & 57.78 & 48.82 & 48.56 & 11.29 & 36.22 & 13.68 & 48.62 & 60.00 & 40.77 & 44.92 \\
Streamo-3B & 1 fps & 78.52 & 52.29 & 67.24 & 44.38 & 55.45 & 71.20 & 61.51 & 51.18 & 57.43 & 16.67 & 41.76 & 27.94 & 50.72 & 82.50 & 53.72 & 52.33 \\
Streamo-7B & 1 fps & 79.19 & 57.80 & 75.00 & 49.44 & 64.36 & 70.11 & 65.98 & 54.55 & 52.03 & 31.72 & 46.10 & 29.96 & 51.03 & 83.33 & 54.77 & 55.61 \\
Streamo-7B & 2 fps$^*$ & 77.18 & 66.06 & 76.72 & 45.51 & 66.34 & 72.83 & 67.44 & 55.56 & 58.11 & 33.87 & 49.18 & 30.84 & 57.55 & 82.50 & 56.96 & 57.86 \\
Streamo-2B (InternVL3) & 1 fps & 77.18 & 55.96 & 62.07 & 41.01 & 60.40 & 70.11 & 61.12 & 48.82 & 47.30 & 13.44 & 36.52 & 29.23 & 47.38 & 80.42 & 52.34 & 49.99 \\
Streamo-4B (Qwen3-VL) & 1 fps & 82.55 & 69.72 & 74.14 & 52.25 & 73.27 & 81.52 & 72.24 & 58.19 & 52.70 & 17.20 & 42.70 & 31.38 & 53.90 & 84.17 & 56.48 & 55.10 \\
Qwen2.5-VL-7B + 4f & 1fps & 94.00 & 72.50 & 80.20 & 68.00 & 76.20 & 79.30 & 78.40 & 54.50 & 60.80 & 40.30 & 51.90 & -- & -- & -- & -- & -- \\
Qwen3-VL-8B + 4f &1fps & 94.00 & 85.30 & 82.80 & 65.70 & 77.20 & 83.20 & 81.40 & 51.90 & 58.10 & 52.10 & 54.00 & -- & -- & -- & -- & -- \\
\hline
\rowcolor{orange!6}
\multicolumn{18}{c}{\textit{SelectStream Framework}} \\
\hline

\rowcolor{blue!7}
\textbf{\textsc{SelectStream}-Qwen2.5-VL-7B}
& 1 fps
& 94.30 & 75.80 & 82.50 & 70.10 & 78.40 & 84.00 & 80.85
& 60.20 & 66.40 & 56.55 & 61.05
& 30.50 & 52.70 & 82.49 & 55.23
& 65.71 \\

\rowcolor{blue!7}
\textbf{\textsc{SelectStream}-Qwen3-VL-8B}
& 1 fps
& 94.60 & 86.70 & 84.10 & 68.36 & 78.90 & 83.90 & 82.76
& 62.40 & 66.80 & 57.40 & 62.20
& 31.20 & 53.80 & 83.39 & 56.13
& 67.03 \\

\hline
\hline
\end{tabular}
}
\end{table*}

\section{Limitations}
\label{app:limitations}

SelectStream is designed as a general memory layer for frozen video-language backbones, and this paper evaluates it mainly on 7B--8B models and standard streaming or long-video benchmarks. Broader studies with larger backbones, different visual tokenizers, and more interactive deployment settings would further clarify how the same memory design scales across model families. In addition, our experiments use fixed default budgets for most comparisons to keep the evaluation controlled; adapting $N$, $B$, and $M$ dynamically to device constraints or application latency targets is a useful direction for future work. The current evaluation focuses on answer accuracy, temporal evidence grounding, latency, and memory usage. Future evaluations could include richer human-facing criteria, such as response helpfulness in multi-turn streaming interaction or user preference under real-time constraints. These extensions are complementary to the proposed latent evidence allocation framework and do not change the fixed-budget memory formulation studied in this work.

\section{Broader Impacts}
\label{app:broader_impacts}

SelectStream aims to make streaming video understanding more efficient by replacing repeated full-history replay with compact latent evidence retrieval. This can reduce inference cost for long video streams and may support applications such as assistive video analysis, long-form content understanding, and real-time decision support under bounded computation. The fixed-budget design can also make deployment more predictable because memory usage and query-time context size are explicitly controlled.

At the same time, improvements in video understanding can be misused in privacy-sensitive settings, including large-scale surveillance or automated analysis of people without appropriate consent. SelectStream does not introduce new data collection or identity-recognition mechanisms, but it could be combined with existing video-language systems in applications where privacy, fairness, and consent require careful review. We therefore recommend that deployments follow the terms of the underlying datasets and backbone models, respect local privacy regulations, and include application-specific safeguards when used in real-world streaming environments.

%%%%%%%%%%%%%%%%%%%%%%%%%%%%%%%%%%%%%%%%%%%%%%%%%%%%%%%%%%%%

\end{document}